\documentclass[lettersize,journal]{IEEEtran}
\usepackage{amsmath,amsfonts}
\usepackage{algorithmic}
\usepackage{algorithm}
\usepackage{array}
\usepackage[caption=false,font=normalsize,labelfont=sf,textfont=sf]{subfig}
\usepackage{textcomp}
\usepackage{stfloats}
\usepackage{url}
\usepackage{verbatim}
\usepackage{graphicx}
\usepackage{cite}
\usepackage{booktabs}
\usepackage{textcomp}
\usepackage{siunitx}
\usepackage{color}
\usepackage{hyperref}
\usepackage{amsmath}
\usepackage{multirow}

\begin{document}

\title{Spiking World Model with Multi-Compartment Neurons for Model-based Reinforcement Learning}

\author{Yinqian Sun, Feifei Zhao, Mingyang Lv, Yi Zeng
\thanks{Manuscript received April 19, 2021; revised August 16, 2021.}
\thanks{Yinqian Sun and Feifei Zhao  is with the Brain-inspired Cognitive Intelligence Lab, Institute of Automation, Chinese Academy of Sciences, Beijing, China (e-mail: sunyinqian2018@ia.ac.cn)}
 \thanks{Mingyang Lv is with the Brain-inspired Cognitive Intelligence Lab, Institute of Automation, Chinese Academy of Sciences, Beijing, China and School of Artificial Intelligence, University of Chinese Academy of Sciences, Beijing, China}
	 \thanks{Yi Zeng is with the Brain-inspired Cognitive Intelligence Lab, Institute of Automation, Chinese Academy of Sciences, Beijing, China, School of Artificial Intelligence, University of Chinese Academy of Sciences, Beijing, China, Key Laboratory of Brain Cognition and Brain-inspired Intelligence Technology, CAS, Shanghai, China and Center for Long-term Artificial Intelligence, Beijing, China.(e-mail: yi.zeng@ia.ac.cn)}
	
	 \thanks{The corresponding author is Yi Zeng.}}
	
\maketitle

\begin{abstract}

Brain-inspired spiking neural networks (SNNs) have garnered significant research attention in algorithm design and perception applications. However, their potential in the decision-making domain, particularly in model-based reinforcement learning, remains underexplored. The difficulty lies in the need for spiking neurons with long-term temporal memory capabilities, as well as network optimization that can integrate and learn information for accurate predictions. The dynamic dendritic information integration mechanism of biological neurons brings us valuable insights for addressing these challenges. In this study, we propose a multi-compartment neuron model capable of nonlinearly integrating information from multiple dendritic sources to dynamically process long sequential inputs. Based on this model, we construct a Spiking World Model (Spiking-WM), to enable model-based deep reinforcement learning (DRL) with SNNs. We evaluated our model using the DeepMind Control Suite, demonstrating that Spiking-WM outperforms existing SNN-based models and achieves performance comparable to artificial neural network (ANN)-based world models employing Gated Recurrent Units (GRUs). Furthermore, we assess the long-term memory capabilities of the proposed model in speech datasets, including SHD, TIMIT, and LibriSpeech 100h, showing that our multi-compartment neuron model surpasses other SNN-based architectures in processing long sequences. Our findings underscore the critical role of dendritic information integration in shaping neuronal function, emphasizing the importance of cooperative dendritic processing in enhancing neural computation.
\end{abstract}


\section{Introduction}

In the burgeoning field of neuromorphic computing, the integration of biologically inspired models into artificial neural networks has paved the way for novel computational frameworks that transcend traditional approaches. Inspired by the dynamic processes of neuronal activity and spike-based signal transmission in the brain, spiking neural networks exhibit energy saving and computational efficiency advantages~\cite{roy2019towards,rathi2023exploring}, and have been effectively utilized across a wide range of applications~\cite{li2022efficient,luo2021siamsnn,sun2022solving,zhou2022spikformer}. Current SNN models predominantly employ Leaky Integrate-and-
Fire (LIF) neurons as a trade-off between biological plausibility and computing simplicity. However, recent advancements in neuroscientific research have revealed the significant role of dendritic structures in neuronal information processing~\cite{larkum1999new,smith2013dendritic,gidon2020dendritic}. The multi-compartment neuron (MCN) model refers to a mathematical framework that treats different parts of a biological neuron such as the axon, soma, and dendrites as independent computational units.

The multi-compartment SNNs captures the morphological characteristics of biological neurons, utilizing dynamic equations to model dendritic information processing, membrane potential propagation, and somatic spike generation~\cite{poirazi2003pyramidal,gidon2020dendritic}. Various implementations of this model, such as the two-compartment neuron model~\cite{urbanczik2014learning,sacramento2018dendritic} and the multi-compartment model with apical and basal dendrites~\cite{richards2019deep}, have demonstrated applications in tasks like MNIST classification and target-based learning~\cite{capone2022burst}. Additionally, the MCN models have been integrated into neuromorphic chips for energy-efficient simulation of neural circuits~\cite{schemmel2017accelerated,yang2019scalable}. 
Recent studies have shown that multi-compartment SNNs outperform LIF-based SNNs in both temporal sequence classification tasks~\cite{zhang2024tc} and neuromorphic datasets~\cite{li2023learning}.

Unlike static image recognition tasks, reinforcement learning necessitates that an agent collect data and refine models through dynamic interactions with the environment. In particular, model-based reinforcement learning methods~\cite{gregor2019shaping,moerland2023model,liu2021sharp} construct a predictive model based on data gathered from the agent's interactions with the environment. This predictive model is subsequently employed to generate synthetic data through simulated interactions with the agent, thereby optimizing the agent's policy. By reducing the number of direct interactions required with the real environment, these methods enhance sample efficiency. For example, the Dreamer model~\cite{hafner2019dream,hafner2023mastering} uses the recurrent state-space model to construct a world model, enabling latent imagination in long-horizon tasks.

Currently, there is a lack of work applying SNNs to model-based reinforcement learning, primarily due to the challenges posed by the need for strong long-term memory capabilities and the ability to effectively learn from the non-stationary distributions of samples in RL environments. 
Biological studies have demonstrated that the complex computational functions of dendrites enhance neuronal pattern separation~\cite{behabadi2012location}, significantly improve the temporal sequence memory of neurons~\cite{hausser2003dendrites,branco2010single}, facilitate rule learning and the formation of long-term memory~\cite{jarsky2005conditional}, and enable neurons to perform more complex perceptual decision-making tasks.

Inspired by the dendritic computational mechanisms of biological neurons, this study presents a multi-compartmental neuron model with sequential nonlinear dendritic integration capabilities. Leveraging this model, we develop a spiking world model to enable a model-based deep reinforcement learning framework using spiking neural networks. The proposed approach achieves performance comparable to conventional RNN-based DRL methods on DeepMind visual control tasks. Moreover, it outperforms existing spiking recurrent neural network models on speech sequence datasets, including SHD, TIMIT, and LibriSpeech 100h. Overall, the main contributions of this work are as follows: 
\begin{itemize}
	\item We propose a multi-compartment neuron model that incorporates dendritic computations, inspired by the structural characteristics of biological neurons. Our spiking MCN model exhibits long-term sequential memory capabilities by leveraging the biophysical dynamics of neuronal compartments, providing enhanced temporal processing capacity for spiking neural networks.

	\item Building upon the multi-compartment neuron framework, we introduce a spiking world model and develop a fully spiking model-based DRL framework. Specifically, we implement an SNN-based Dreamer model, offering a biologically plausible alternative for model-based reinforcement learning in spiking neural networks.

	\item We evaluate the proposed Spiking-WM and MCN models on DeepMind visual control tasks and sequential speech datasets, including SHD, TIMIT, and LibriSpeech 100h. Experimental results demonstrate that our models surpass existing state-of-the-art spiking recurrent neural networks with the same parameter budget, establishing a new benchmark for SNN-based sequence modeling and control tasks. 
\end{itemize}

\begin{figure*}
	\centering
	\includegraphics[width=0.95\linewidth]{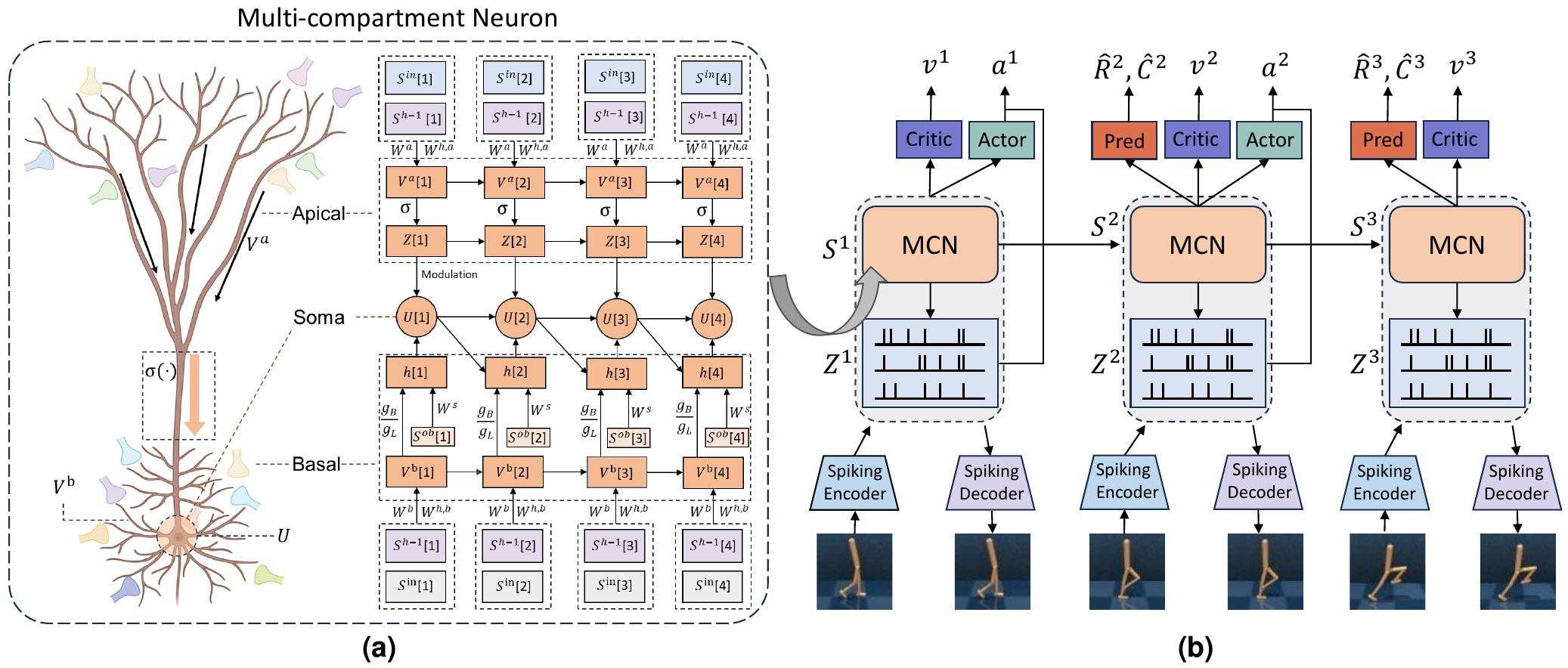} 
	\caption{The spiking world model based on MCN. \textbf{(a)} Biological structure of pyramidal neurons (left) and the multi-compartment neuron model (right). \textbf{(b)} The architecture of spiking world model.}
	\label{Fig:Spike-WM-all}
\end{figure*}

\section{Related Work}
Most spiking neural network research utilizes point-neuron models to represent biological neurons, such as the widely used LIF~\cite{gerstner2002spiking} neuron model, which computing process relying solely on the somatic membrane potential. The ALIF~\cite{yin2021accurate} model proposes an adaptive spiking recurrent neural network to process time-domain datasets classification. The improved recurrent spiking  neuron \cite{ponghiran2022spiking} modified LIF neurons to enhance their inherent recurrence for sequential learning, demonstrating that the improved SNN architecture achieves comparable accuracy to LSTMs on speech recognition tasks.

Point neuron models lack structural detail in neuronal modeling and are insufficient for accurately capturing the complex information integration and long-sequence processing capabilities of biological neurons. Consequently, there is growing research interest in computational modeling of dendritic processes and the development of multi-compartmental neuron models. A two-compartment neuron model with distinct soma and dendrite components is introduced by ~\cite{urbanczik2014learning,sacramento2018dendritic}, which fine-tunes the dendritic synapse weights to facilitate dendritic predictions of somatic spike firing patterns. Additionally, \cite{richards2019deep} proposed a three-compartment neuron model that captures the computational processes occurring in apical dendrites, basal dendrites, and the soma.  The TC-LIF~\cite{zhang2024tc} built a two-compartment spiking neuron model to enhance long-term sequential modeling in SNNs. Target-based learning~\cite{lansdell2019learning,capone2022burst} leverages the apical dendrite to receive the plasticity signal, enabling local modification of neuron weights.

However, the aforementioned multi-compartmental neuron models provide only a simplified representation of dendritic computations, with most implementations incorporating a basic neuronal structure containing a single dendrite. Additionally, they largely overlook the nonlinear interactions and memory storage mechanisms involved in dendrite-to-soma signal transmission. To address these limitations, this study considers the distinct computational roles of dendrites at different locations within biological neurons, such as the apical and basal dendrites of pyramidal neurons. Furthermore, we model the cooperative computational processes that occur during signal transmission across dendrites at different locations.

Currently, research on brain-inspired SNNs is primarily focused on visual information processing areas. The application of SNNs, particularly those with deep structures, in complex decision-making fields remains relatively limited. One major challenge for implementing RL algorithms using SNNs is the difficulty in training and converging the model, owing to the dynamic nature of RL environments. 

Some studies have applied spiking neural networks to reinforcement learning tasks. Using the ANN-to-SNN conversion method, the convolutional layers and fully connected layers in the DQN network are converted to SNN formulation applied to the Atari Breakout game~\cite{patel2019improved,tan2021strategy}. 

In the area of continuous control tasks, SNNs based on R-STDP have been applied to intelligent vehicle lane-keeping tasks~\cite{bing2020indirect} and training hexapod robots to walk~\cite{lele2020learning}. PopSAN~\cite{tang2021deep} uses a population coding method with spiking neurons to implement an SNN-based reinforcement learning Actor-Critic algorithm, which is used for continuous control tasks in MuJoCo simulation environment. A two-module spiking neural network model is introduced that leverages biologically inspired mechanisms, such as "dreaming" (offline learning in a model-based simulated environment) and "planning" (online learning), to enhance reinforcement learning efficiency without requiring the detailed storage of past experiences~\cite{capone2022towards}. The MCS-FQF \cite{SUN2025106898} presents a brain-inspired deep distributional reinforcement learning algorithm based on multi-compartment neuron model SNN and population coding method, which significantly enhances computational power and performance in Atari games.

Nevertheless, existing research on reinforcement learning with spiking neural networks has primarily focused on implementing model-free RL methods, while the application of SNNs in model-based RL remains largely unexplored. In this study, we develop a spiking world model based on a multi-compartmental neuron model with nonlinear dendritic computations, enabling the application of SNNs in model-based deep reinforcement learning.

\section{Preliminary and Methodology}

\subsection{Model-based Reinforcement Learning}

Reinforcement learning can be formed as a Markov Decision Process (MDP), which can be described by element matrix: $<\mathcal{S},\mathcal{A},\mathcal{P}^a_{ss'},\mathcal{R}^a_{ss'},\gamma>$. At discrete time step $t\in [1;T]$, the agent takes action with probability $a_t\sim p(a_t|s_t)$ with $a_t\in \mathcal{A}; s_t \in \mathcal{S}$. The state transiting to $s_{t+1}\sim p(s_{t+1}\in \mathcal{S}|s_t,a_t)$ when agent taking action $a_t$ at state $s_t$ with probability $\mathcal{P}^{a_t}_{s_ts_{t+1}}$, and the agent receives the reward $r_t\sim \mathcal{R}^{a_t}_{s_ts_{t+1}}$ generated by the unknown environment. Typically, the main objective of reinforcement learning is to maximize the expected cumulative reward $E_{p}(\sum_{t=1}^T\gamma \cdot r_t)$.

Model-based reinforcement learning represents the world environment with latent dynamic model, using probability functions $q(s_t|s_{t-1}, a_{t-1})$ to fit the state transition process $\mathcal{P}^{a_t}_{s_ts_{t+1}}$. Similarly, the rewards from the environment can also be predicted by $q(r_t|s_t)$. Compared to model-free reinforcement learning, model-based reinforcement learning can reduce the number of interactions with the environment by learning policies through the world model. This alleviates the trial-and-error cost of reinforcement learning algorithms and increases the feasibility of deploying these algorithms in real-world environments.

\subsection{Multi-compartment Neuron Model}

The spiking neuron model simulates the propagation of electrical potentials and the firing processes of biological neurons. Taking the LIF neuron as an example, which is designed primarily for balancing the intricate complexity of biological neuron details with computational tractability, and is widely used in the construction of spiking neural networks and brain simulations. The membrane potential dynamics of the LIF neuron are described by Equation ~\ref{EQ:LIF}, and the parameter $\tau$ is the somatic decay time parameter. When the membrane potential $U_t$ exceeds a threshold $V_{th} $, the neuron fires a spike $S_t$ and the membrane potential is reset $U_t \leftarrow V_{res}; U_t>V_{th}$. 

\begin{equation}
	\tau\frac{dU_t}{dt} = -U_t+ x_t;  \quad  U_t<V_{th}
	\label{EQ:LIF}
\end{equation}

\begin{equation}
	S_t =	\left\{\begin{array}{l}
		1 \quad U_t > V_{th} \\
		0  \quad U_t < V_{th}
	\end{array}\right.
	\label{EQ:spike}
\end{equation}

Although the membrane potential dynamic process of LIF spiking neurons has the inherent ability to process sequential information, it only considers the passive updating of membrane potential and spike firing process. This approach ignores the morphological and structural characteristics of neurons and models neurons as point-like structures without taking into account the local computational capabilities and functional enhancements provided by different parts of biological neurons, such as dendrites and soma.

Biological neurons can utilize different parts of their dendrites to receive input signals from various sources, enabling local independent computations. In contrast, the LIF neuron model only considers somatic computation. This paper proposes a multi-compartment neuron model that incorporates the collaborative computation of multiple dendrites and the soma. Specifically, pyramidal neurons, which are widely present in the hippocampus and neocortex, are modeled as computational units consisting of three parts: apical dendrites, basal dendrites, and the soma, shown as Figure~\ref{Fig:Spike-WM-all}(a). The dynamic processing of the proposed MCN model is as follows:
\begin{equation}
	\tau\frac{dU_t}{dt}=-U_t + \frac{g_B}{g_L}(V^b_t-U_t)+\frac{g_A}{g_L}(V^a_t-U_t)
	\label{EQ:mc}
\end{equation}
where $V^b$ and $V^a$ are the basal and apical dendritic membrane potentials, respectively. The $g_B$, and $g_A$ are conductance parameters for  basal and apical dendrites, and $g_L$ is membrane leaky conductance. Same as LIF model, the MCN neuron fires a spike when the somatic potential $U_t$ exceeds the threshold.

\subsection{Sequential Nonlinear Dendritic Integration for MCN}
Spiking neurons, by modeling the dynamic biophysical processes of biological neurons, possess the inherent ability to process sequential information due to the mathematical form of their dynamic differential equations. In this work, we propose a method for sequential information processing based on multi-compartment neurons and apply it to model-based deep reinforcement learning. Inspired by the brain mechanism where information from different parts of a biological neuron's dendrites has varying modulatory effects on the somatic membrane potential, we formulate the MCN model with local membrane potential computation processes of basal and apical dendrites as

\begin{equation}
	\left\{\begin{array}{l}
		\tau_b\frac{dV_t^b}{dt}=-V_t^b + x_t^b \\
		x_t^b=W^bS_t^{in}+W^{h,b}S_{t}^{h-1}
	\end{array}\right.
	\label{EQ:mc_basal}
\end{equation}

\begin{equation}
	\left\{\begin{array}{l}
		\tau_a\frac{dV_t^a}{dt}=-V_t^a + x_t^a \\
		x_t^a=W^aS_t^{in}+W^{h,a}S_{t}^{h-1}
	\end{array}\right.
	\label{EQ:mc_apical}
\end{equation}

\begin{equation}
	\tau\frac{dU_t}{dt}=-U_t+\sigma(V_t^a)(\frac{g_B}{g_L}(V_t^b-U_t)+W^sS_{t}^{in})    
	\label{EQ:mc_soma}
\end{equation}
where the $V_t^b$ and $V_t^a$ is basal and apical dendritic potentials, respectively. Both change dynamically based on the external sources of spike signals $S_t^{in}$ and  internal signals $S_{t}^h$ representing hidden states. The basal and apical dendrites are equipped with synaptic weights $[W^b, W^{h,b}]$ and $[W^a, W^{h,a}]$, which are adjustable parameters providing the dendrites with local learning capabilities.    

This work primarily explores the potential recurrent memory function inherent in pyramidal neurons. One important source of these internal signals $S_{t}^h$ can be the recurrent and feedback loop connections that are abundantly present in the hippocampus and neocortex. Similar to the multi-compartment neuron model described in Equation ~\ref{EQ:mc}, the computational results of the dendrites are integrated at the soma in the form of membrane potential. However, in this work, we explore the nonlinear modulatory effects of apical dendrites on the somatic potential $U_t$, where the potential of the apical dendrites $V_t^a$ determines the contribution of the information transmitted by the basal dendrites $V_t^b$ and the soma's input information $S_{t}^{in}$ which is processed by  $W^s$ the learnable somatic synapse parameter.

\subsection{Spiking World Model}

The membrane potential dynamics of the spiking neuron model make it suitable for processing sequential signals. Based on the proposed MCN model, we built a spiking State Space Model (SSM) as in Dreamer work~\cite{hafner2023mastering}. Furthermore, we use LIF neurons to construct the model's encoder, decoder, and policy network, thereby implementing a fully spiking world model. 
In this work, we handle continuous control tasks with visual inputs using LIF neurons to construct a spiking convolutional neural network (SCNN), which extracts features from image inputs and performs spiking encoding as Equation~\ref{EQ:scnn}. The $O[t]$ is an observation input at time $t$ and the SCNN encodes it to spikes $S^{ob}[t]$.
A fully connected LIF-based SNN model is used to integrate the previous latent state $Z^{h-1}[t]$ and action $A^{h-1}[t]$ to generate sequential model spike input $S^{in}[t]$.
\begin{small}
\begin{equation}
	S^{ob}[t] = SCNN(O[t]); S^{in}[t]=SNN(Z^{h-1}[t],A^{h-1}[t])
	\label{EQ:scnn}
\end{equation}
\end{small}

We used the Euler method to numerically expand the Equation~(\ref{EQ:mc_basal}--\ref{EQ:mc_soma}) MCN formulation, proposing the following sequential model:

\begin{small}
\begin{equation}
	V^b[t]= V^b[t-1] + (W^bS^{in}[t]+W^{h,b}S^{h-1}[t]-V^b[t-1])/\tau_b
	\label{EQ:basal_t}
\end{equation}
\end{small}

\begin{small}
\begin{equation}
	V^a[t]= V^a[t-1] + (W^aS^{in}[t]+W^{h,a}S^{h-1}[t]-V^a[t-1])/\tau_a
	\label{EQ:apical_t}
\end{equation}
\end{small}

\begin{small}
\begin{equation}
	h[t]=\frac{g_B}{g_L}(V^b[t]-U[t-1]) + W^sS^{ob}[t]; \  z[t]=\sigma(V^a[t])
	\label{EQ:h_t}
\end{equation}
\end{small}

\begin{equation}
	U[t]=U[t-1]+(z[t]\odot h[t]-U[t-1])/\tau
	\label{EQ:soma_t}
\end{equation}

\begin{equation}
	S^{h}[t]=\Theta(U[t]-V_{th})
	\label{EQ:spike_t}
\end{equation}
where $\Theta(\cdot)$ is the Heaviside step function. When firing a spike, the MCN resets its membrane potential as $U[t]=U[t](1-S^{h}[t])$. We use the Sigmoid funtion with parameter $\beta$ as the gating function for the apical dendritic membrane potential with 
\begin{equation}
	\sigma(x) = \frac{1}{1+e^{-\beta x}}
	\label{EQ:sigmoid}
\end{equation}

The state spiking encoding $S^h[t]$ of the multi-compartment neuron model, along with the dynamic updates of the neuron's dendritic and somatic membrane potentials, endows it with long-short term memory capabilities.

The observation encoding and sequence model output are used to generate the model state representation $Z^h[t]$: 
\begin{equation}
	S^h[t] = MCN(S^{in}[t], S^{h-1}[t]) 
\end{equation}
\begin{equation}
	Z^h[t] \sim q_\theta(Z^h[t]|S^{ob}[t], S^{h}[t])) 
\end{equation}

Similar to the Dreamer model, we use the SNNs to output distributional state parameters instead of the direct values of the state representation. The model state $Z^h[t]$ and the spiking information output $S^{h}[t]$ from the MCN are used to predict the reward $\hat{R}[t]$ and the continuous signal $\hat{C}[t]$ for each episode. The decoder uses a spiking deconvolutional network to obtain predictions of observation $\hat{O}[t]$ from the model's latent state. The predictive outputs of the world model are used to select actions and evaluate states.

\begin{equation}
   \hat{Z}^h[t] \sim p_\theta(\hat{Z}^h[t]|S^{h}[t]) 
\end{equation}

\begin{equation}
	 \hat{R}^h[t] \sim p_\theta(\hat{R}^h[t]|S^{h}[t],Z^h[t])  
\end{equation}

\begin{equation}
	\hat{C}^h[t] \sim p_\theta(\hat{C}^h[t]|S^{h}[t],Z^h[t]) 
\end{equation}

\begin{equation}
   \hat{O}[t] \sim p_\theta(\hat{O}[t]|S^{h}[t],Z^h[t])
\end{equation}

 Figure~\ref{Fig:Spike-WM-all}(b) shows the spiking world model and action selection process. We use an SNN model to construct the policy model, which includes both actor and critic networks. The actor network integrates the spiking sequence representation of the world model state $[S^{h}_t, Z^{h}_t]$ at spiking time step $t$ to determine the action selection $a^h$ at the $h$-th step within the environmental context. The critic network generates environmental state values $v^h$ based on the spiking sequence representation of the model as an estimate of the expected returns obtainable from the current state. 
\begin{equation}
	\begin{array}{ll}
		\text{Actor:} & a^h \sim SNN(S^{h}_t, Z^{h}_t) \\
		\text{Critic:} & v^h \sim SNN(S^{h}_t, Z^{h}_t)  \\
	\end{array}
\end{equation}

\begin{figure*}[htbp]
	\centering
	\includegraphics[width=1.0\linewidth]{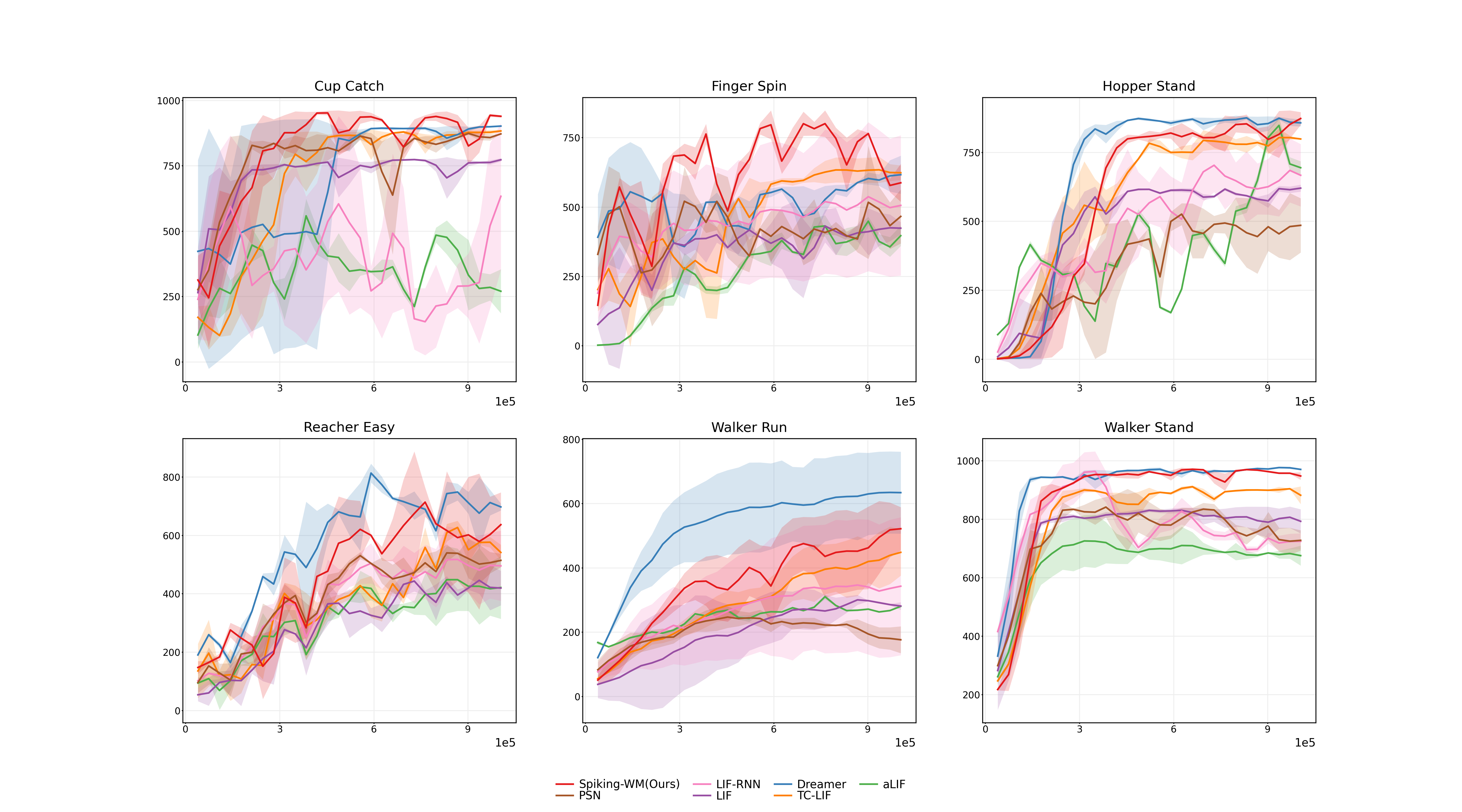} 
	\caption{Evaluation scores for DeepMind visual control experiments.}
	\label{Fig:dmc-res-plot}
\end{figure*}

We train the actor and critic using the bootstrapped method. Within the predicted horizon of length $H$, the $\lambda$-returns are calculated based on the current reward predictions $\hat{R}^{1:H}$ and environmental condition predictions $\hat{C}^{1:H}$, as shown by Equation~\ref{EQ:lam_ret}. As in Dreamer method, SNN based actor and critic are trained by discrete regression and maximum returns.

\begin{small}
\begin{equation}
	\mathcal{R}^{\lambda, h} = \hat{R}^h + \gamma \hat{C}^h[t]((1-\lambda)v^{h+1}+\lambda\mathcal{R}^{\lambda, h+1}) \quad \lambda\mathcal{R}^{\lambda, H}\doteq  v^{H}
	\label{EQ:lam_ret}
\end{equation}
\end{small}

The loss function of the world model consists of the model prediction loss $\mathbb{L}_{pred}(\theta)$ and the posterior estimation loss $\mathbb{L}_{model}(\theta)$, as shown in Equation~\ref{EQ:loss}. The objective of the model prediction loss  $\mathbb{L}_{pred}(\theta)$  is to optimize the maximum likelihood of the decoder, reward predictor, and continue predictor under the true observations, rewards, and continue condition values of the environment. The model dynamics prediction loss $\mathbb{L}_{model}(\theta)$ aims to improve the sequence model MCN's prediction of the next state by minimizing the KL divergence between the prior predictions and the posterior representations using Bayesian estimation methods.
\begin{equation}
	\mathbb{L}_{model}(\theta) = E_{q_{\theta}}[\sum_{h=1}^{H}(\mathbb{L}_{pred}(\theta)+\mathbb{L}_{model}(\theta))]
	\label{EQ:loss}
\end{equation}

\begin{equation}
	\mathbb{L}_{pred}(\theta) = \sum_{x^h} -lnp_\theta(x^h|,S^{h}[t],Z^h[t]) \quad x^h\in\{O^h,r^h,c^h\} 
	\label{EQ:pred_loss}
\end{equation}
	
\begin{equation}
	\mathbb{L}(\theta) = KL[q_\theta(Z^h|S^{ob}[t], S^{h}[t]) || p_\theta(\hat{Z}^h[t]|S^{h}[t])]
	\label{EQ:model_loss}
\end{equation}

Considering that the long-short term memory processes of SNNs and MCN involve the temporal integration of spike sequences, we optimize the SNNs' parameters by using the Spatio-temporal Backpropagation (STBP) method~\cite{SUN2025106898}. At the discontinuous points of spike firing, we use the a surrogate function $g'(x)$ as the gradient of spike firing moments:

\begin{equation}
	\frac{\partial S^h[t]}{\partial U[t]} \longleftarrow g'(x)(U[t]-V_{th})
	\label{EQ:arctan}
\end{equation}

\begin{equation}
	g'(x) =
	\begin{cases}
		0, & |x| > \frac{1}{\alpha} \\\\
		-\alpha^2|x|+\alpha, & |x| \leq \frac{1}{\alpha}
	\end{cases}
\end{equation}

\section{Experiments}
To evaluate the performance of the proposed spiking world model, we conducted experiments on the DeepMind Visual Control Suite. Additionally, to investigate the long-sequence memory capability of the MCN model, we tested the performance of different neurons on the speech datasets SHD, TIMIT, and LibriSpeech 100h. Furthermore, to assess the impact of MCN parameters on model performance, we employed grid search and adaptive parameter learning methods to analyze the task performance and long-sequence memory capacity of the Spiking-WM model under different dendritic parameter configurations of the MCN.

\subsection{DeepMind Visual Control Experiments}

\begin{table*}[!h]
	\centering
	\begin{tabular}{l r r r r r r r}	
		\toprule
		\textbf{Task}               & Dreamer(GRU)~\cite{hafner2023mastering}      & LIF~\cite{cramer2020heidelberg}       & aLIF~\cite{yin2021accurate}     & PSN~\cite{fang2024parallel}        & LIF-RNN~\cite{ponghiran2022spiking}   & TC-LIF~\cite{zhang2024tc}    & Spiking-WM(Ours)  \\
		\midrule
		\textbf{Parameters (M)}     &  19.10            & 19.11     & 18.85    & 18.84      & 19.11     &  18.84         & 18.84  \\
		\textbf{Env Steps (M)}      & 1.0               & 1.0       & 1.0      & 1.0        & 1.0       & 1.0            & 1.0   \\
		\midrule
		Acrobot Swingup             & \textbf{219.1}             & 104.0     & 79.1     & 95.8     & 125.5     & 106.5     & 113.7  \\ 
		Cartpole Balance            & 970.7             & 663.0     & 502.4    & 788.6    & 957.7     & 965.3     & \textbf{973.3} \\ 
		Cartpole Balance Sparse     & 991.2             & 991.5     & 748.3    & 832.6    & 718.2     & \textbf{992.4}     & 989.6  \\ 
		Cartpole Swingup            & 810.1             & 774.1     & 614.0    & 575.9    & 772.6     & 667.7     & \textbf{791.0} \\ 
		Cartpole Swingup Sparse     & 480.2             & 261.3     & 217.1    & 240.6    & 211.1     & 265.5     & \textbf{420.1} \\ 
		Cheetah Run                 & \textbf{720.1}             & 515.5     & 558.1    & 283.9    & 510.8     & 517.8     & 577.2 \\ 
		Cup Catch                   & 958.2             & 768.2     & 494.2    & 915.8    & 523.0     & 938.7     & \textbf{960.3} \\ 
		Finger Spin                 & 637.3             & 416.2     & 433.5    & 472.4    & 509.3     & 657.8     & \textbf{682.0} \\ 
		Finger Turn Easy            & \textbf{763.5}             & 489.2     & 542.9    & 592.7    & 706.1     & 688.5     & \textbf{727.5} \\ 
		Finger Turn Hard            & 397.2             & 193.6     & 201.3    & 289.1    & 269.4     & 369.1     & \textbf{406.3} \\ 
		Hopper Hop                  & \textbf{373.4}             & 104.9     & 145.7    & 114.4    & 215.6     & 205.4     & 234.9 \\ 
		Hopper Stand                & 913.6             & 650.6     & 704.1    & 493.1    & 715.1     & 864.2     & \textbf{905.3} \\ 
		Pendulum Swingup            & \textbf{800.4}             & 660.9     & 665.0    & 556.4    & 669.2     & 746.2     & \textbf{796.5} \\ 
		Quadruped Run               & \textbf{690.4}             & 440.7     & 425.3    & 303.9    & 485.9     & 372.1     & \textbf{515.3} \\ 
		Quadruped Walk              & \textbf{387.7}             & 259.7     & 260.6    & 280.3    & 320.3     & 302.2     & \textbf{350.7} \\ 
		Reacher Easy                & \textbf{736.2}             & 403.4     & 422.0    & 511.5    & 503.1     & 613.2     & \textbf{701.3} \\ 
		Walker Run                  & \textbf{705.5}             & 320.4     & 325.8    & 264.1    & 352.2     & 460.8     & 541.6 \\ 
		Walker Stand                & \textbf{965.9}             & 799.8     & 724.6    & 790.2    & 784.6     & 916.7     & \textbf{960.5} \\ 
		Walker Walk                 & \textbf{955.1}             & 795.2     & 871.5    & 706.0    & 924.0     & 916.1     & \textbf{938.5} \\ 
		\midrule
		\textbf{Mean}               & \textbf{709.2}             & 505.9     & 470.2    & 479.3    & 540.7     & 608.7     & \textbf{662.4} \\ 
		\textbf{GRU\%}              & -                 & 67.0      & 63.2     & 64.3     & 73.8      & 81.9      & \textbf{90.4} \\ 
		\bottomrule		 
	\end{tabular}
	\caption{Scores for DMC visual inputs control experiments at 1M frames.}
	\label{Tab:dmc_res}
\end{table*}

\begin{figure*}
	\centering
	\includegraphics[width=0.9\linewidth]{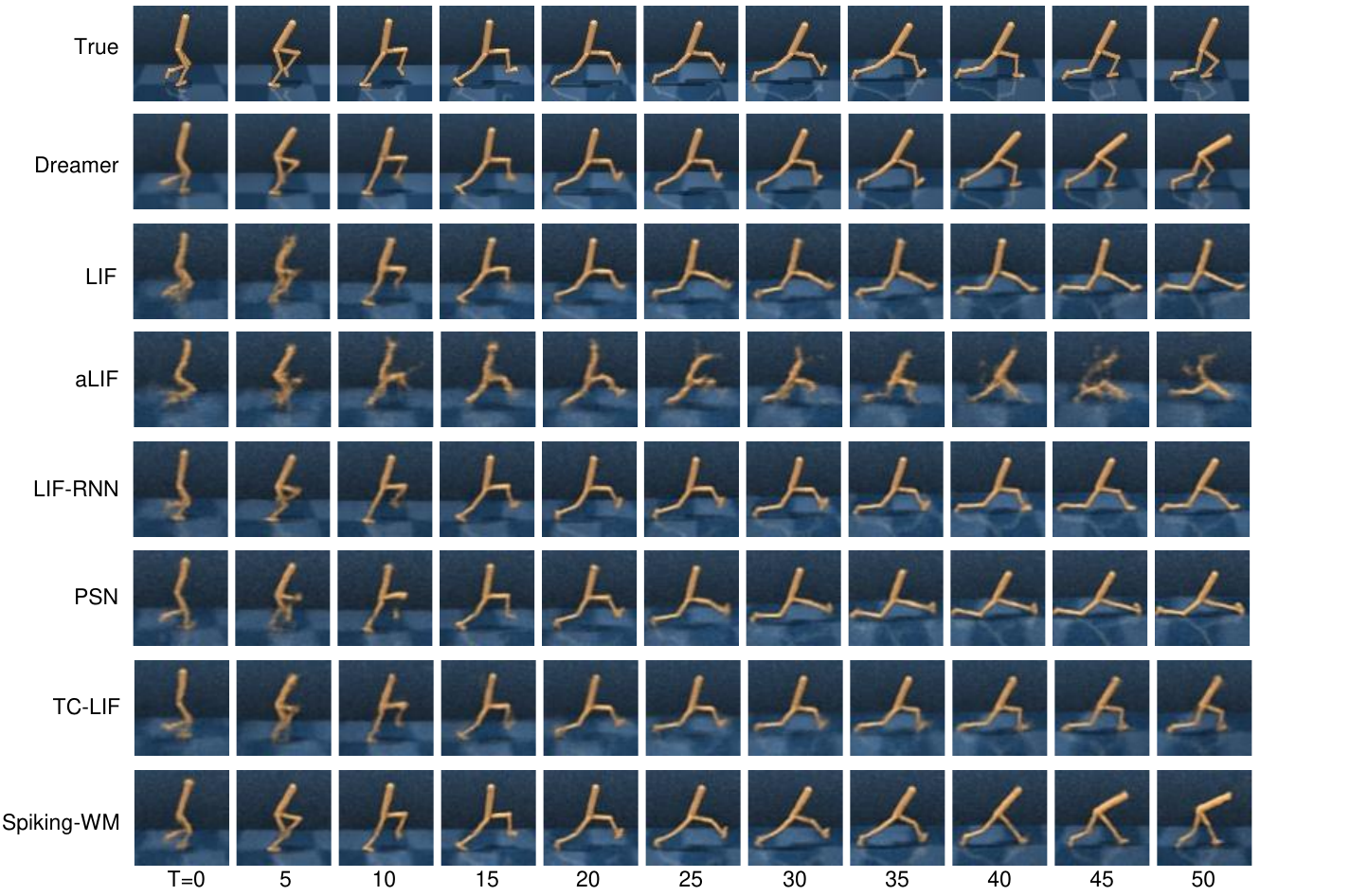} 
	\caption{Comparison of visual state predictions for different neuron models.}
	\label{Fig:prdict}
\end{figure*}

\begin{table}[]
	\centering
	\resizebox{.95\columnwidth}{!}{
		\begin{tabular}{l l l l l l}
			\toprule[2pt]
			Parameter & Value & Description & Parameter & Value & Description \\
			\midrule[1pt]
			$\tau$  & 2.0  & Soma decay parameter &$g_L$ & 1.0   & Leaky conductance  \\
			$V_{th}$  & 1.0  & Threshold potential &$g_B$ & 1.0  & Basal conductance  \\
			$V_{reset}$ & 0.0  & Reset potential  &$\beta$ & 1.0  & Gating parameter\\
			$T_s$  & 8 ms & Simulation Times        &$lr $ & $1.0e^{-4}$ &  Learning rate  \\
			$\tau_A$ & 2.0  & Apical decay parameter     \\
			$\tau_B$ & 2.0  & Basal decay parameter &  & &\\
			\bottomrule[2pt]
	\end{tabular}}
	\caption{Parameter settings for experiments.}
	\label{Tab:parameter}
\end{table}

We conducted experiments on visual input control tasks in DeepMind Control (DMC) Suite, comparing the performance of MCN-based spiking world model across 19 continuous control tasks. For comparison, we evaluated the capabilities of the traditional GRU-based world model algorithm Dreamer~\cite{hafner2023mastering}. In addition, to demonstrate the critical role of the MCN model in the spiking world model and its advantages over other point-neuron models and simple two-compartment neuron models, we constructed equivalent spiking world models based on several alternatives. These included the aLIF~\cite{yin2021accurate} model, formed by augmenting the baseline LIF~\cite{cramer2020heidelberg} neuron for processing long-sequence spiking data, the LIF-RNN~\cite{ponghiran2022spiking} model, which enhances internal recurrence, the parallelized spiking neuron (PSN)~\cite{fang2024parallel} and the two-compartment neuron (TC-LIF)~\cite{zhang2024tc} model. Experiments were conducted on the DeepMind Visual Control suite using identical model parameters and training settings. Detailed model and experimental parameter settings are shown in Table~\ref{Tab:parameter}.

DMC tasks based on visual input require the model to accurately represent the features of the current observation images and make correct control actions based on visual perception. In the experiment, we use a 4-layer spiking convolutional network to process image observations from the environment, encoding the image observation into the spiking sequence space. All compared models use the same SCNN model structure and parameter settings to process visual inputs and employ the same SNN-based Actor-Critic policy learning.

In the DMC experiments, the sequence prediction model spiking-WM consists of 512 multi-compartment neurons. For the sake of fairness in comparison, we increased the number of neurons and the dimensionality of weights in the other baseline models to ensure consistency in the total number of parameters. Detailed parameter statistics and experimental results are provided in Table~\ref{Tab:dmc_res}. The decoder uses a 4-layer spiking transposed convolutional network, and the predictor is a fully connected LIF neuron SNN. 

The final evaluation results were obtained after training for one million (1M) frames. The results in Figure~\ref{Fig:dmc-res-plot} and Table~\ref{Tab:dmc_res} demonstrate that our proposed Spiking-WM model outperforms other spiking neuron baselines in terms of learning speed and final scores, achieving results comparable to the ANN-based Dreamer model. On the average score across 19 tasks, the Spiking-WM model achieves 90.4$\%$ of the performance of the ANN-based Dreamer(GRU) model, surpassing the TC-LIF model, which reached only 81.9$\%$ of Dreamer(GRU) scores, and outperforming point-neuron models like LIF and its derivative models by 31$\%$\~{}38$\%$.

Additionally, the Spiking-WM model exceeds the performance of Dreamer(GRU) in four tasks: Cartpole Balance, Cup Catch, Finger Spin, Finger Turn Hard. In contrast, other spiking neuron-based world models generally do not outperform Dreamer(GRU) in terms of task performance, with the exception of TC-LIF on the Cartpole Balance Sparse task. These results demonstrate the advantages of our proposed MCN over existing simple neuron models in implementing the Spiking-WM model and highlight the potential of biologically plausible brain-inspired spiking neural networks for model-based reinforcement learning algorithms.

Using the trained model, we sampled the predictions of world models with different neurons for the first 50 steps of the Walker Run task, as shown in Figure~\ref{Fig:prdict}. Here, T=0 represents the input state, and T=5-50 represents the model's Decoder prediction outputs. The results show that our proposed MCN-based world model, Spiking-WM, achieves better predictive capabilities, leading to higher final task scores compared to other neural world models.

\subsection{Sequential Speech Data Experiments}
\begin{table}[]
	\centering
    \tabcolsep=4pt
	\caption{Comparison of Different Neuron Types on Long-Sequence Datasets. In the “Neuron Type” category, “FF" denotes neurons that primarily compute information in a feedforward manner, while “Rec" represents neurons with an enhanced design for recurrent information processing within the neuron itself.}
        \begin{tabular}{c|ccccccccccc}
			\toprule
			Dataset            & Method                                    & Neuron Type    & Params (M)   & Acc(\%)\\
			\midrule
			\multirow{7}{*}{SHD}  
			                   & GRU                                   	&  Rec           & 3.5              & 89.92 \\
							   & LIF~\cite{cramer2020heidelberg}       	&  FF            & 2.4              & 72.06 \\
							   & aLIF~\cite{yin2021accurate}           	&  Rec           & 2.5              & 88.64 \\
                                  & PSN~\cite{fang2024parallel}            &  FF            & 2.0              & 85.86 \\
							   & LIF-RNN~\cite{ponghiran2022spiking}   	&  Rec           & 2.1              & 80.34 \\
							   & TC-LIF~\cite{zhang2024tc}             	&  FF            & 2.1              & 89.45 \\
							   & Ours                      &  Rec           & 2.5              & \textbf{89.57} \\
			\midrule
			\multirow{7}{*}{TIMIT}  
							   & GRU              					   	& Rec            & 14.1             & 83.75 \\
							   & LIF~\cite{cramer2020heidelberg}        & FF             & 13.9             & 78.35 \\
							   & aLIF~\cite{yin2021accurate}            & Rec            & 11.4             & 82.10 \\
							   & PSN~\cite{fang2024parallel}            & FF             & 11.8             & 80.05 \\
							   & LIF-RNN~\cite{ponghiran2022spiking}    & Rec            & 19.6             & 81.28 \\
							   & TC-LIF~\cite{zhang2024tc}           	& FF             & 11.9             & 80.90 \\
							   & Spiking-WM(Ours)    					& Rec            & 14.0             & \textbf{83.01} \\
							\midrule
							\multirow{7}{*}{LibriSpeech}  
							   & GRU              						& Rec            & 5.1              & 90.22 \\
							   & LIF~\cite{cramer2020heidelberg}        & FF             & 3.9              & 80.16 \\
							\multirow{5}{*}{100h} 
							   & aLIF~\cite{yin2021accurate}            & Rec            & 3.7              & 87.08 \\
							   & PSN~\cite{fang2024parallel}            & FF             & 2.6              & 87.56 \\
							   & LIF-RNN~\cite{ponghiran2022spiking}    & Rec            & 3.8              & 88.25 \\
						       & TC-LIF~\cite{zhang2024tc}           	& FF             & 3.9              & 88.59 \\
							   & Spiking-WM(Ours)    					& Rec            & 3.9              & \textbf{88.77} \\
			\bottomrule
		\end{tabular}
		\label{Tab:seq_res}
	\end{table}

To validate the potential of the proposed multi-compartment neuron model in handling sequence models, we conducted tests on the speech datasets SHD, TIMIT and LibriSpeech, the resuts are shown in Table~\ref{Tab:seq_res}.
Different neural models share an identical network architecture and training parameter configurations when applied to the same dataset. For the SHD dataset, the network adopts a fully connected structure comprising two hidden layers, each containing 128 neurons, with the STSC~\cite{yu2022stsc} spike processing method implemented at the input layer, which has been demonstrated to significantly enhance the performance of SNNs in handling spike information~\cite{shen2023eventmix,li2024parallel}.

We conducted comparative experiments on the TIMIT and LibriSpeech 100h datasets using a two-layer neural network with the same recurrent connection and experimental settings as same in paper~\cite{ponghiran2022spiking}. The TIMIT dataset is a widely used benchmark for speech recognition, containing phonetically diverse sentences spoken by multiple speakers, while LibriSpeech 100h is a subset of the larger LibriSpeech corpus, consisting of approximately 100 hours of transcribed English speech derived from audiobook recordings. The network architecture follows a two-layer recurrent structure, maintaining the same parameter configurations as in the referenced study LIF-RNN~\cite{ponghiran2022spiking}. 
Experimental results demonstrate that our proposed spiking-WM model outperforms other spiking neuron models on the SHD, TIMIT and LibriSpeech 100h datasets, achieving performance comparable to that of GRU, which is consistent with the results of the DeepMind visual control experiments.

\begin{figure*}
	\centering
	\includegraphics[width=0.9\linewidth]{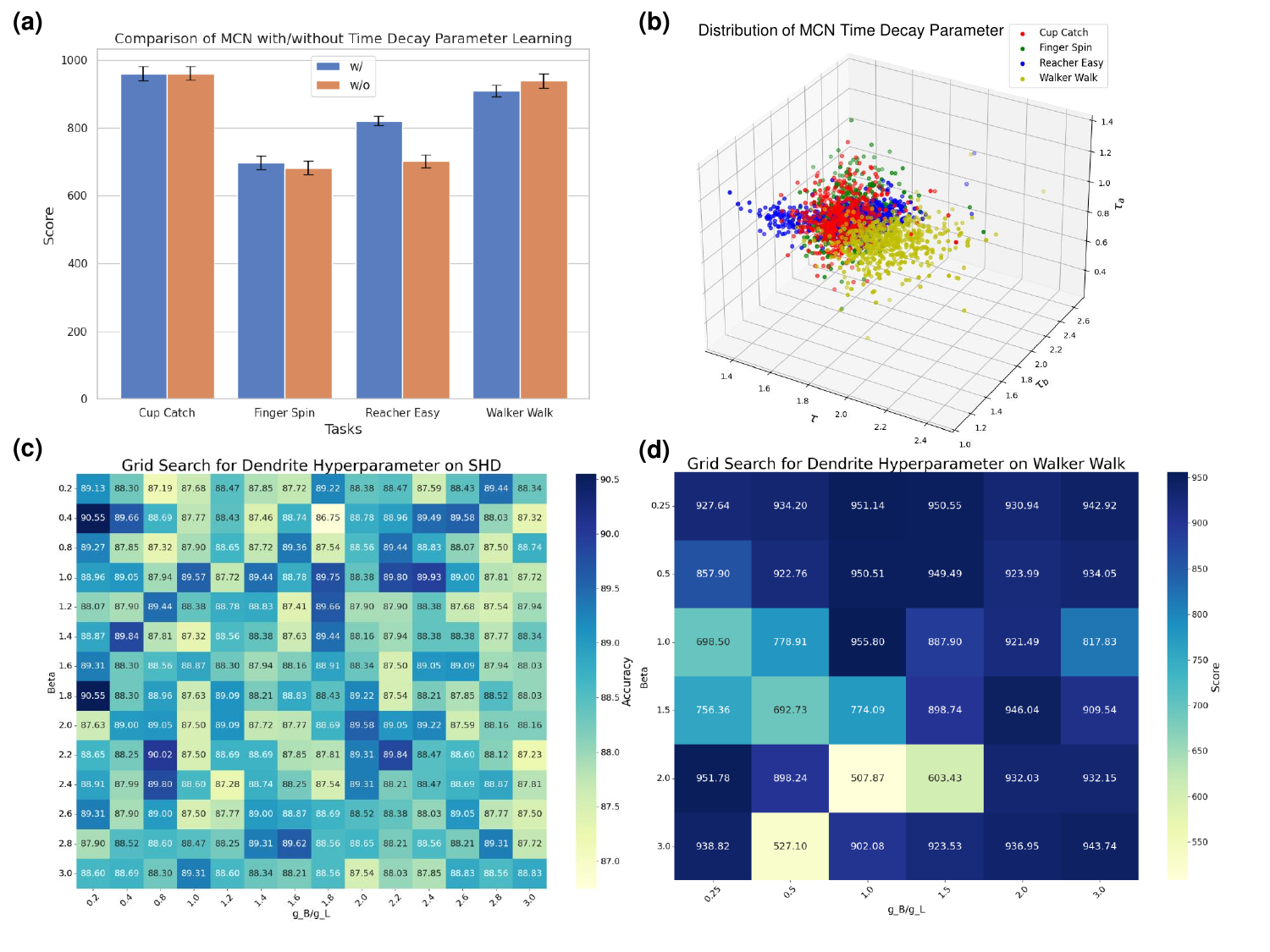} 
	\caption{Experimental results on the impact of MCN parameters on model performance. \textbf{(a)} Comparison results of experiments on MCN with and without learnable time decay parameters. \textbf{(b)} The distibution of soma, basal dendrite and apical dendrite time parameters after learning. \textbf{(c)} Grid search of basal dendrite conductance parameters and apical dendrite gate parameter on SHD dataset. \textbf{(d)} Grid search of basal dendrite conductance parameters and apical dendrite gate parameter on Walker Walk.}
	\label{Fig:hyper_grid}
\end{figure*}

\subsection{Analysis of MCN Parameters}
Compared to the LIF neuron model, the multi-compartment neuron model introduces more parameters through the refinement of dendritic structures, thus providing more adjustable components for the neuron model. In our experiment, we investigated the impact of MCN parameters on model performance. For the membrane time constants of the soma ($\tau$) and dendritic membrane ($\tau_a,\tau_b$), we applied the approach from PLIF~\cite{fang2021incorporating}, treating them as learnable parameters and conducting experiments on the DMC task. The results are shown in Figure~\ref{Fig:hyper_grid}(a)(b). In the four experimental tasks, the learnable membrane potential decay parameters in the Finger Spin and Reacher Easy tasks led to performance improvements. However, in the Cup Catch task, the learnable decay parameters did not yield higher scores compared to the fixed decay parameters, and even resulted in a negative performance gain in the Walker Walk task. 

The statistical distribution of the trained membrane potential decay parameters for dendrites and the soma is presented in Figure~\ref{Fig:hyper_grid}(b). Although the final distributions of the optimized membrane potential decay parameters exhibit slight variations across the four tasks, a significant portion of the distribution regions remain overlapping. Combined with the results in Figure~\ref{Fig:hyper_grid}(a), this suggests that adjustments to different membrane time constants in the MCN model do not substantially impact overall model performance. Instead, the key contribution of the MCN model lies in the incorporation of dendritic computational processes.

Furthermore, we analyzed the impact of the basal dendritic conductance parameter $\frac{g_B}{g_L}$ and the apical dendritic nonlinear function parameter $\beta$ on the MCN model. Using a grid search approach, we tested the model performance with different values of $\frac{g_B}{g_L}$ and $\beta$ on the SHD dataset and the DMC Walker Walk task. The results are shown in Figure~\ref{Fig:hyper_grid}(c)(d).

The parameter search results indicate that the synergistic interaction between apical and basal dendrites significantly impacts the performance of the MCN model. On the SHD dataset, higher accuracy is observed in the upper-left region of Figure~\ref{Fig:hyper_grid}(c) ($\frac{g_B}{g_L} \leq 1.8$, $\beta \leq 1.8$). In the Walker Walk task, $\frac{g_B}{g_L}$ and $\beta$ exhibit a more distinct adaptation relationship, with high scores concentrated in the upper-right region of Figure~\ref{Fig:hyper_grid}(d). Moreover, better performance is more likely to be achieved when $\beta$ falls within the range of 0.5 to 1.0 and $\frac{g_B}{g_L}$ within the range of 1.0 to 1.5.

\begin{figure*}
	\centering
	\includegraphics[width=0.9\linewidth]{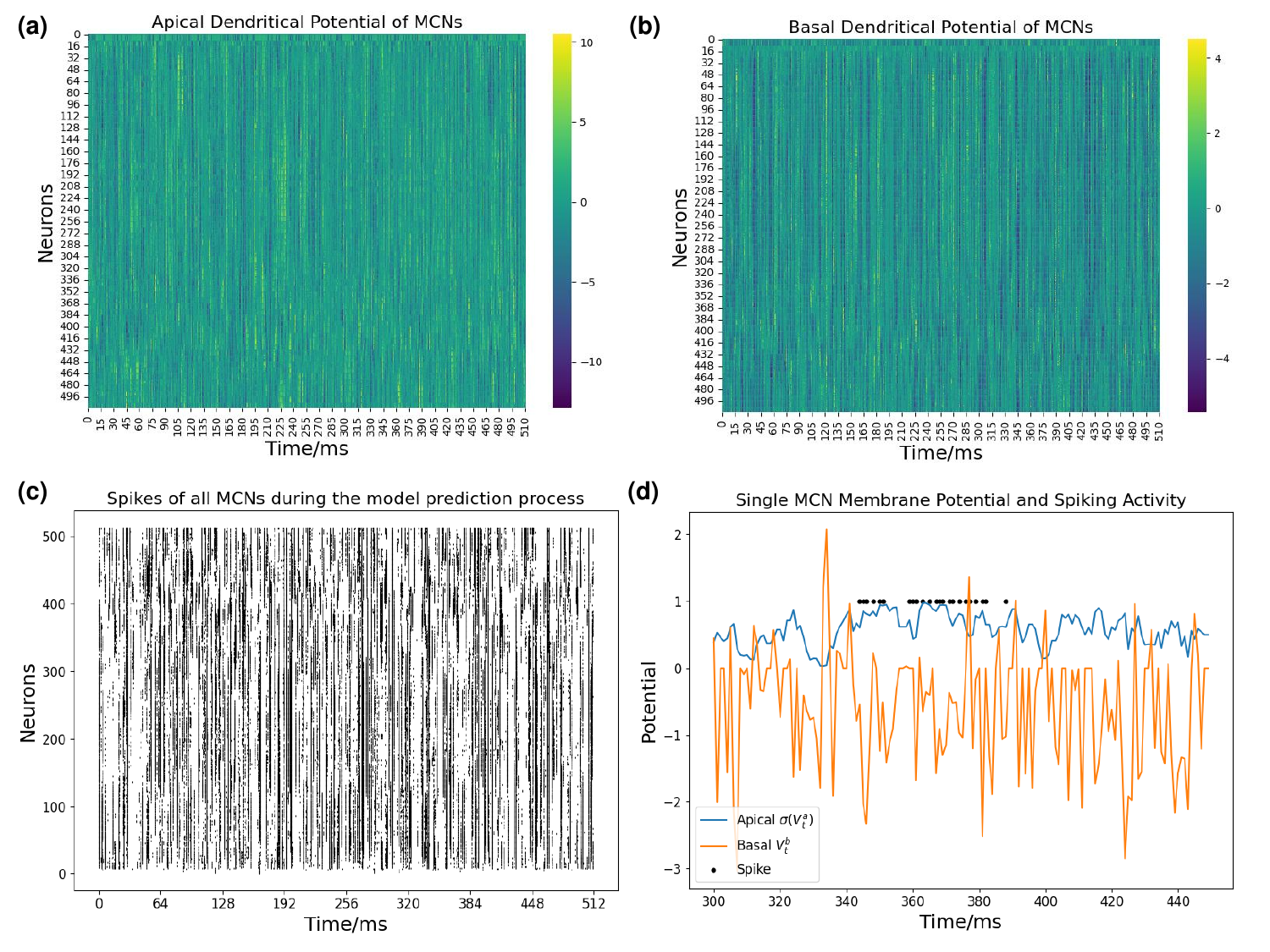} 
	\caption{Statistics of dendritic membrane potential and spike firing activity of all multi-compartment neurons at task Walker Run predition process. \textbf{(a)} The apical dendritical potential of MCNs. \textbf{(b)} The basal dendritical potential of MCNs. \textbf{(c)} The spike sequential of MCNs during the model prediction process. \textbf{(d)} Details of the dendritic membrane potential and spike firing of the 200th neuron.}
	\label{Fig:spike_hist}
\end{figure*}

\subsection{Analysis of MCN Spiking Activity}

We further examined the membrane potentials and spike output of MCN neurons during the model prediction phase of the Spiking-WM in the DMC Walker Run task, as shown in Figure~\ref{Fig:spike_hist}. During this process, the dendritic membrane potentials of MCN neurons exhibit synchronized activity over time, as illustrated in Figure~\ref{Fig:spike_hist}(a) and Figure~\ref{Fig:spike_hist}(b), which leads to a burst firing pattern of the neuron, as shown in Figure~\ref{Fig:spike_hist}(c). Furthermore, the frequency of apical dendritic membrane potential activity is higher than that of the basal dendrites, as indicated by the larger areas of alternating light and dark regions in the statistical plot.

The Figure~\ref{Fig:spike_hist}(d) presents the recorded dendritic membrane potential, somatic membrane potential, and spike firing of a single neuron (Neuron 200) within the 300 ms to 450 ms interval. To facilitate observation of the regulatory effect of apical dendrites on neuronal activity, we have included the gating signal $z[t]=\sigma(V^a[t])$ (the blue line) of the apical dendrite in the figure. The neuron exhibits significant spike firing activity during the 350 ms to 380 ms period, during which the gating signal of the apical dendritic membrane potential remains within the range of 0.5$\sim$1.0 for an extended duration. At this time, the acpial dendritic membrane potential is greater than zero, corresponding to the bright regions in Figure~\ref{Fig:spike_hist}(a). In contrast, although the basal dendritic membrane potential exhibits large fluctuations at other recorded time points, the corresponding apical dendrite remains in a scaled inhibition state (with the gating signal mostly below 0.5), preventing the neuron from generating spike firing. These findings highlight the modulatory effect of apical dendritic activity on neuronal spike output in the proposed MCN model.

\section{Conclusion}

This work introduces a multi-compartment neuron-based spiking world model, demonstrating its effectiveness in model-based SNN deep reinforcement learning and long-term memory tasks. Through comprehensive evaluations on DeepMind Control suits and large-scale speech datasets, our approach consistently outperforms conventional SNN models and achieves competitive performance relative to ANN-based world models. Moreover, our analysis of neuronal membrane potentials, spiking behavior, and dendritic parameter optimization underscores the significance of dendritic information processing. We show that coordinated dendritic activity plays a crucial role in shaping neuronal responses and improving overall model performance. These findings provide new insights into the functional importance of dendritic computation in SNNs, paving the way for more biologically inspired and efficient neural architectures in future research.

\subsection{Data availability}
The model of this research is one of the core and part of BrainCog Embot~\cite{embot_braincog}. BrainCog Embot is an Embodied AI platform under the Brain-inspired Cognitive Intelligence Engine (BrainCog) framework, which is an open-source Brain-inspired AI platform based on Spiking Neural Network.

\section*{Acknowledgments}
 This study is supported by the Strategic Priority Research Program of the Chinese Academy of Sciences (Grant No. XDB1010302), the funding from Institute of Automation, Chinese Academy of Sciences (Grant No. E411230101),  Postdoctoral Fellowship Program of CPSF (Grant No.GZC20232994) the National Natural Science Foundation of China (Grant No. 62406325).

\ifCLASSOPTIONcaptionsoff
\newpage
\fi
%

\bibliographystyle{IEEEtran}
\bibliography{Refs}

\newpage

%
%
%
%

\vfill

\end{document}